\documentclass{article}

\PassOptionsToPackage{numbers}{natbib}

\usepackage[preprint]{neurips_2025}

\usepackage[utf8]{inputenc} % allow utf-8 input
\usepackage[T1]{fontenc}    % use 8-bit T1 fonts
\usepackage{hyperref}       % hyperlinks
\usepackage{url}            % simple URL typesetting
\usepackage{booktabs}       % professional-quality tables
\usepackage{amsfonts}       % blackboard math symbols
\usepackage{nicefrac}       % compact symbols for 1/2, etc.
\usepackage{microtype}      % microtypography
\usepackage{xcolor}         % colors
\usepackage{amsmath}
\usepackage{enumitem}
\usepackage{graphicx}
\usepackage{makecell}

\usepackage{array}

\usepackage[ruled]{algorithm2e}

\usepackage{algpseudocode}
\usepackage{caption}
\usepackage{float}
\usepackage{booktabs}
\usepackage{multirow}

\usepackage{graphicx}
\usepackage{subfigure}

\usepackage{tcolorbox}

\title{LoHoVLA: A Unified Vision-Language-Action Model \\for Long-Horizon Embodied Tasks}

\author{%
  Yi Yang \\
  Fudan University\\
  \texttt{21307130076@m.fudan.edu.cn} \\
  \And
  Jiaxuan Sun \\
  ShanghaiTech University \\
  \texttt{sunjx2022@shanghaitech.edu.cn} \\
  \And
  Siqi Kou \\
  Shanghai Jiao Tong University \\
  \texttt{happy-karry@sjtu.edu.cn} \\
  \And
  Yihan Wang \\
  Fudan University \\
  \texttt{23301170011@m.fudan.edu.cn} \\
  \And
  Zhijie Deng\thanks{Corresponding author.} \\
  Shanghai Jiao Tong University \\
  \texttt{zhijied@sjtu.edu.cn} \\
}

\begin{document}

\maketitle

\begin{abstract}
% Below is the abstract of a NeurIPS paper. Polish the writing to meet the academic style, improve the spelling, grammar, clarity, concision and overall readability. Simplify lengthy or cumbersome sentences. When necessary, rewrite the whole sentence. Firstly, you should provide the polished paragraph. Secondly, you should list all your modification and explain the reasons to do so in markdown table.
Real-world embodied agents face long-horizon tasks, characterized by high-level goals demanding multi-step solutions beyond single actions. Successfully navigating these requires both high-level task planning (i.e., decomposing goals into sub-tasks) and low-level motion control (i.e., generating precise robot actions). While existing vision language action (VLA) models and hierarchical architectures offer potential in embodied tasks, the former often falter in planning, and the latter can suffer from coordination issues, both hampering performance. 
We introduce a new unified VLA framework for long-horizon tasks, dubbed LoHoVLA, to overcome these limitations. %. As , LoHoVLA integrates task planning and motion control 
% LoHoVLA employs a large pretrained Vision Language Model (VLM) backbone, with dedicated language and action de-tokenizer for sub-task planning and motor command prediction, respectively. This shared backbone facilitates learning common representations for both task types, improving generalization.
LoHoVLA leverages a large pretrained vision language model (VLM) as the backbone to jointly generate language and action tokens for sub-task generation and robot action prediction, respectively. 
This shared representation promotes better generalization across tasks.
Additionally, LoHoVLA embraces a hierarchical closed-loop control mechanism to mitigate errors originating from both high-level planning and low-level control. 
To train LoHoVLA, we introduce LoHoSet, a dataset built on the Ravens simulator, containing 20 long-horizon tasks, each with 1,000 expert demonstrations composed of visual observations, linguistic goals, sub-tasks, and robot actions. 
% LoHoVLA adopts a mixed training strategy that jointly optimizes sub-task generation and robot action prediction. 
Experimental results show that LoHoVLA significantly surpasses both hierarchical and standard VLA approaches on long-horizon embodied tasks in the Ravens simulator. %, showcasing superior reasoning, planning, and generalization. 
These findings underscore the promise of unified architectures for advancing generalizable embodied intelligence. 
% Real-world embodied agents must tackle long-horizon tasks, characterized by high-level goals that inherently require multi-step solutions rather than single actions. 
% This requires both high-level task planning and low-level motion control: the former decomposes overarching goals into atomic sub-tasks, while the latter generates precise motor commands to execute them. 
% Although prior approaches using Vision–Language–Action (VLA) models and hierarchical architectures have shown potential in this case, VLA models often struggle with planning, and hierarchical architectures tend to face challenges in coordination, leading to suboptimal performance. 
% To address these issues, we propose LoHoVLA, a unified VLA model that integrates high-level task planning and low-level motion control within a single framework. 
% LoHoVLA leverages a large pretrained Vision Language Model (VLM) backbone, equipped with a language head for sub-task planning and an action head for motor command prediction. 
% The shared backbone learns shared representations across both task types and enables improved generalization across scenarios. 
% We evaluate LoHoVLA on the LoHoRavens benchmark, where it outperforms both hierarchical architectures and standard VLA models, demonstrating superior reasoning, planning, and generalization capabilities. 
% These results highlight the potential of unified architectures in advancing real-world embodied intelligence.
\end{abstract}

\section{Introduction}
\label{sec:intro}
% I am writing a NeurIPS paper. Please help me polish the sentences to make them more concise, formal, logical, and natural, like those written by a native speaker. At the same time, you should replace any inaccurate or unnatural words.

% Below is a paragraph from a NeurIPS paper. Polish the writing to meet the academic style, improve the spelling, grammar, clarity, concision and overall readability. Simplify lengthy or cumbersome sentences. When necessary, rewrite the whole sentence. Firstly, you should provide the polished paragraph. Secondly, you should list all your modification and explain the reasons to do so in markdown table.

Embodied agents operating in real world are required to handle tasks that are long-horizon, compositional, and dynamically changing~\cite{liu2024aligning,duan2022survey,wang2024karma,wu2023embodied}. 
Unlike short-horizon tasks~\cite{yu2020meta, james2020rlbench, zeng2021transporter, shridhar2022cliport, zhu2020robosuite, jiang2022vima}, the long-horizon ones involve high-level goals that cannot be achieved in a single action. 
Agents must do reasoning, execute movements, and adapt to failures or changes in the environment. 
This necessitates both high-level task planning and low-level motion control, with the former decomposing the overall goal into atomic tasks and the latter generating accurate robot actions for execution.

Vision language action (VLA) models have emerged as a dominant paradigm for embodied agents~\cite{brohan2023rt, kim2024openvla, team2024octo}. 
They usually use large-scale pretrained vision language models (VLMs) as backbones and are fine-tuned on robotic demonstrations to map visual and linguistic inputs to executable robot actions. 
While VLA models effectively extract key information from observations and instructions, 
% \zj{our experiments demonstrate that they face challenges in planning and reasoning, revise: they cannot/fall short in xxx, don't mention our experiment results},
they fall short in effective planning and reasoning, suffering from subpar performance on long-horizon tasks.

Seminal studies on long-horizon embodied tasks usually employ hierarchical architectures~\cite{zhao2023erra, yang2024guiding, ahn2022can, liang2023code, zhang2023lohoravens, huang2022inner}, including a high-level VLM-based planner to reason about sub-task instructions and a low-level VLA-based controller to convert these instructions into robot actions. 
While this modular structure provides flexibility, it frequently results in suboptimal coordination and limited generalization~\cite{intelligence2025pi_, zhang2024vlabench}. 
These limitations underscore the need for unified, end-to-end architectures that combine high-level and low-level inference in a single framework.

To bridge the gap, we introduce \textbf{LoHoVLA}, a unified \textbf{V}ision–\textbf{L}anguage–\textbf{A}ction model for \textbf{Lo}ng-\textbf{Ho}rizon embodied tasks that integrates both high-level task planning and low-level motion control. 
LoHoVLA first infers linguistic sub-tasks from the input observations and the specified high-level goal, then uses the inferred sub-tasks as contextual guidance for action prediction. 
% If the current sub-task remains incomplete\zj{why sub-task can be incomplete? confusing.}, the model directly predicts actions without re-inferring sub-tasks.
The robot executes the predicted actions to interact with and modify the environment. New observations are subsequently captured to infer the next sub-tasks and the following actions.

We build LoHoVLA upon a large pretrained VLM backbone to leverage its extensive world knowledge and reasoning capabilities~\cite{cheang2024gr, li2024towards}. 
We augment the original language head to generate both linguistic sub-tasks and discrete action tokens.
This shared backbone enables learning of generalizable representations across both planning and control.
To further enhance robustness, we introduce a hierarchical closed-loop control mechanism: If the execution of a sub-task fails for times exceeding a predefined threshold, the system re-plans the sub-task; otherwise, it updates only the actions based on the new environment state.

% \zj{Considering the scarcity of data annotated with fine-grained reasonings and actions, we construct a synthetic dataset, UniRaven,s based on the Ravens simulator for the training of LoHoVLA. Specifically, xxx} 
Considering the scarcity of long-horizon demonstrations annotated with fine-grained sub-tasks and actions, we synthesize the dataset \textbf{LoHoSet} to train LoHoVLA. Specifically, LoHoSet is developed based on the Ravens robot simulator and comprises 20 long-horizon embodied tasks. Each task includes 1,000 expert demonstrations featuring visual observations, linguistic goals, sub-tasks, and actions. The LoHoVLA trained on the dataset can significantly outperform both the hierarchical baseline and a vanilla VLA baseline on both seen and unseen tasks, showing strong reasoning and planning capabilities as well as strong generalization. These findings underscore the potential of unified architectures for real-world embodied intelligence.

\section{Related Work}

\subsection{Vision-Language-Action Models}
Vision language action (VLA) models have become a central paradigm in robot learning by bridging perception, language understanding, and control. These models typically repurpose large-scale vision language models (VLMs)~\cite{alayrac2022flamingo,guo2024sam2point,chen2023pali,karamcheti2024prismatic} pretrained on internet-scale datasets for downstream robotic tasks. Such pretrained VLMs offer rich semantic priors and strong generalization across modalities, making them attractive backbones for visuomotor policies. Recent works~\cite{driess2023palm,kim2024openvla,o2024open} fine-tune these backbones on robot demonstration datasets to map image and language inputs to motor actions, showing promising results in generalizing to novel objects, instructions, and environments. 

Despite these advances, most existing VLA frameworks are limited in their ability to perform multi-step reasoning or structured task decomposition, which are critical for executing complex or long-horizon tasks. To address this, some approaches introduce external planners~\cite{yang2024guiding,erdogan2025plan,zhou2024isr} or use action experts~\cite{wen2025dexvla} to handle sub-task planing, though this modularity often incurs coordination overhead and reduces system robustness. Furthermore, while pretrained VLMs offer strong grounding capabilities, their integration into control pipelines remains challenging due to the mismatch between vision-language pretraining objectives and the fine-grained demands of robotic control. Our work contributes to the growing effort to enhance VLA models by exploring architectural designs that support structured reasoning and low-level control within a cohesive framework.

\subsection{Long-Horizon Embodied Task Planning}
Effective execution of long-horizon tasks in embodied agents necessitates advanced planning capabilities that can decompose high-level goals into coherent sequences of sub-tasks. Recent approaches~\cite{ahn2022can,belkhale2024rt,chen2024automating,cheng2024navila,dai2024racer,hu2023look,li2023interactive,li2025hamster,nasiriany2024pivot,shi2024yell} have leveraged large language models (LLMs) to enhance such planning processes. These methods typically employ LLMs to generate structured plans from natural language instructions, which are then executed by specialized low-level controllers. To improve adaptability, some frameworks~\cite{chen2024automating,dai2024racer,nasiriany2024pivot,shi2024yell} incorporate re-planning mechanisms that allow the agent to adjust its strategy in response to environmental changes, thereby enhancing robustness and generalization across diverse tasks and settings. Beyond textual planning, there is a growing interest in integrating chain-of-thought (CoT) reasoning~\cite{wei2022chain} into embodied systems. This paradigm~\cite{zhao2025cot,zawalski2024robotic,wen2024diffusion} involves generating intermediate reasoning steps that bridge the gap between high-level instructions and low-level actions. By conditioning action prediction on these subgoals, agents can achieve more structured and interpretable behaviors. However, existing approaches often suffer from limitations such as reliance on modular architectures that separate high-level planning and low-level control, leading to suboptimal coordination and limited generalization. Our proposed LoHoVLA model integrates both high-level task planning and low-level motion control within a single unified framework, leveraging a shared vision-language backbone to improve reasoning, planning, and generalization capabilities in long-horizon tasks.

\section{LoHoVLA}
\label{sec:meth}
% Below is a paragraph from a NeurIPS paper. Polish the writing to meet the academic style, improve the spelling, grammar, clarity, concision and overall readability. Simplify lengthy or cumbersome sentences. When necessary, rewrite the whole sentence. Firstly, you should provide the polished paragraph. Secondly, you should list all your modification and explain the reasons to do so in markdown table.

This section elaborates on LoHoVLA, starting with various modeling configurations, followed by a description of the LoHoSet dataset for training LoHoVLA. We then present the model architecture and outline the training strategies.

\begin{figure*}[t]
\centering
\includegraphics[width=1.0\textwidth]{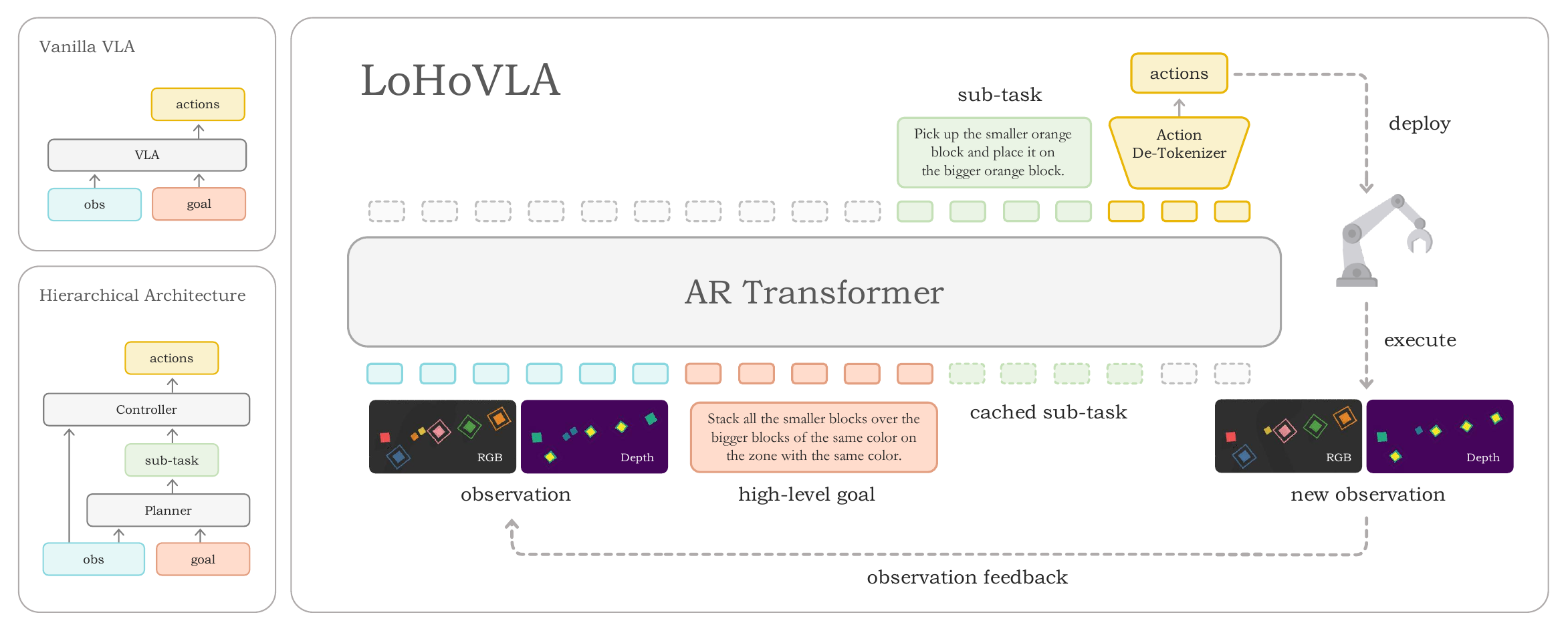}
\caption{
\textit{Left top:} Vanilla VLA directly maps high-level goals and observations to actions.
\textit{Left bottom:} The hierarchical architecture separates planning and execution— the planner infers sub-tasks, and the controller executes them.
\textit{Right:} LoHoVLA integrates high-level task planning and low-level motion control into a unified model. It uses an auto-regressive (AR) Transformer as its backbone and employs a hierarchical closed-loop control mechanism.
}
\label{fig:architecture}
\end{figure*}

\subsection{Modeling Configurations}
\label{3.1}

% The action $\mathbf{a} = (\mathcal{T}_{\text{pick}}, \mathcal{T}_{\text{place}})$ defines the end-effector poses for the pick and place operations, respectively. We focus on tabletop manipulation tasks, where $\mathcal{T}_{\text{pick}}, \mathcal{T}_{\text{place}} \in \textbf{SE}(2)$. The visual observation $\mathbf{o}_t = (\mathcal{I}_{\text{color}}, \mathcal{I}_{\text{depth}})$ includes both RGB and depth top-down orthographic reconstructions of the scene.

We address the problem of learning a %goal-conditioned 
policy $\pi_\theta$ that produces robot actions $\mathbf{a}_t$ based on a visual observation $\mathbf{o}_t$ and a high-level language goal $g$.
The visual observation $\mathbf{o}_t$ consists of images of the scene captured by the robot's cameras.
The goal $g$ defines a high-level instruction for a long-horizon task (e.g., “Clean the desk”), which implicitly incurs the sequential execution of multiple sub-tasks $[\hat{g}_1, \hat{g}_2, \cdots, \hat{g}_N]$ 
(e.g., \textit{``Put the pen in the pen holder''} $\rightarrow$ \textit{``Close the laptop''} $\rightarrow$ \textit{``Put the book on the bookshelf''} $\rightarrow$ etc.). 
For simplicity, we assume that each sub-task can be completed within a single time step.
The action $\mathbf{a}_t$ represents low-level robot actions, such as specifying the Cartesian position of the robot’s end-effector. These commands are executed by the robot to interact with the environment. New observations $\mathbf{o}_{t+1}$ are then captured to determine subsequent actions $\mathbf{a}_{t+1}$.

One typical trajectory for this task can be represented as 
$\{g, \gamma_1, \gamma_2, \cdots ,\gamma_N\}$,
where $g$ denotes the overall goal and 
% $\gamma_i = \{\hat{g}_i, \mathbf{o}_{1:T}^{(i)}, \mathbf{a}_{1:T}^{(i)}\}$ 
% $\gamma_i = \{\hat{g}_i, {(\mathbf{o}_t^{(i)}, \mathbf{a}_t^{(i)})}^{T}_{t=1}\}$ 
$\gamma_t = (\mathbf{o}_t, \hat{g}_t, \mathbf{a}_t)$ 
denotes a triplet of observation, sub-task, and robot action at time step $t$.
% The policy $\pi_\theta$ can be tuned on a collection of such demonstrations are used to supervise the training of t.
During inference, the policy is only provided with the overall goal and does not have access to the sub-tasks. Consequently, it must rely on implicit or explicit planning to infer the sub-tasks.

\textbf{Vanilla VLA} The vanilla VLA model (Figure \ref{fig:architecture}, left top) is limited to action generation and cannot produce language outputs.
Thus, in fact, it implicitly performs sub-task planning, with only a resultant action $\mathbf{a}_t$ yielded based on the high-level goal $g$ and the current observation $\mathbf{o}_t$:
\begin{equation}
\label{eq2}
    \pi_\theta(\mathbf{o}_t, g) \rightarrow \mathbf{a}_t.
\end{equation}
% In our experiments, we trained the vanilla VLA model on the LoHoSet dataset without sub-task labels to evaluate the effectiveness of such implicit planning.

\textbf{The Hierarchical Architecture} The implicit sub-task of VLA lacks interpretability and reliability. To address this, the hierarchical architecture (Figure \ref{fig:architecture}, left bottom) proposes to use an external high-level planner to explicitly infer the next atomic sub-task based on the current observation and the high-level goal. 
Then the low-level controller generates actions to execute this sub-task:
\begin{equation}
\label{eq3}
    \pi_\theta^\text{planner} (\mathbf{o}_t, g) \rightarrow \hat{g}_t, \; \pi_\theta^\text{controller} (\mathbf{o}_t, \hat{g}_t) \rightarrow \mathbf{a}_t.
\end{equation}
% We use LoHoRavens as the hierarchical architecture baseline for comparison with our method.

\textbf{LoHoVLA} Rather than employing disjoint modules that may suffer from suboptimal coordination and modeling redundancy, we advocate a unified paradigm that integrates high-level task planning and low-level motion control into a model (Figure \ref{fig:architecture} right). In formal, there is:
\begin{equation}
%     \pi_\theta(\mathbf{a}_t, \hat{g}_t \mid \mathbf{o}_t, g) =  \begin{cases} \pi_\theta(\mathbf{a}_t \mid \mathbf{o}_t, \hat{g}_t) \cdot \pi_\theta(\hat{g}_t \mid \mathbf{o}_t, g), & \text{if } \beta(\hat{g}_{t-1}, \mathbf{o}_t) > 0 \quad\\ \pi_\theta(\mathbf{a}_t \mid \mathbf{o}_t, \hat{g}_{t}), \quad \text{with } \hat{g}_t = \hat{g}_{t-1}, & \text{else}.\end{cases}
\label{eq4}
    \pi_\theta(\mathbf{a}_t, \hat{g}_t \mid \mathbf{o}_t, g) =   \pi_\theta(\mathbf{a}_t \mid \mathbf{o}_t, g, \hat{g}_t) \cdot \pi_\theta(\hat{g}_t \mid \mathbf{o}_t, g).
\end{equation}
% Here, $\beta$ denotes the environment reward. If $\beta > 0$, indicating that the current sub-task has been completed, LoHoVLA first infers the next atomic sub-task. It then uses the inferred sub-task as contextual guidance to predict the robot's action. If the sub-task is incomplete ($\beta = 0$), the model reuses the cached sub-task to directly predict the next action. As a result, high-level reasoning is triggered less frequently than low-level action inference, distinguishing LoHoVLA from conventional embodied chain-of-thought methods.
As the equation implies, LoHoVLA first infers the next atomic sub-task and then uses it as contextual guidance to predict the robot’s action. High-level task planning corresponds to modeling $\pi_\theta(\hat{g}_t \mid \mathbf{o}_t, g)$, while low-level motion control corresponds to modeling $\pi_\theta(\mathbf{a}_t \mid \mathbf{o}_t, g, \hat{g}_t)$, both distributions are represented within a single unified model.
\subsection{A Synthetic Dataset for Long-horizon Embodied Tasks: LoHoSet}
\label{3.2}
% Below is a paragraph from a NeurIPS paper. Polish the writing to meet the academic style, improve the spelling, grammar, clarity, concision and overall readability. Simplify lengthy or cumbersome sentences. When necessary, rewrite the whole sentence. Firstly, you should provide the polished paragraph. Secondly, you should list all your modification and explain the reasons to do so in markdown table.

\begin{figure*}[t]
\centering
% \hspace*{-0.16\textwidth}
\includegraphics[width=1.0\textwidth]{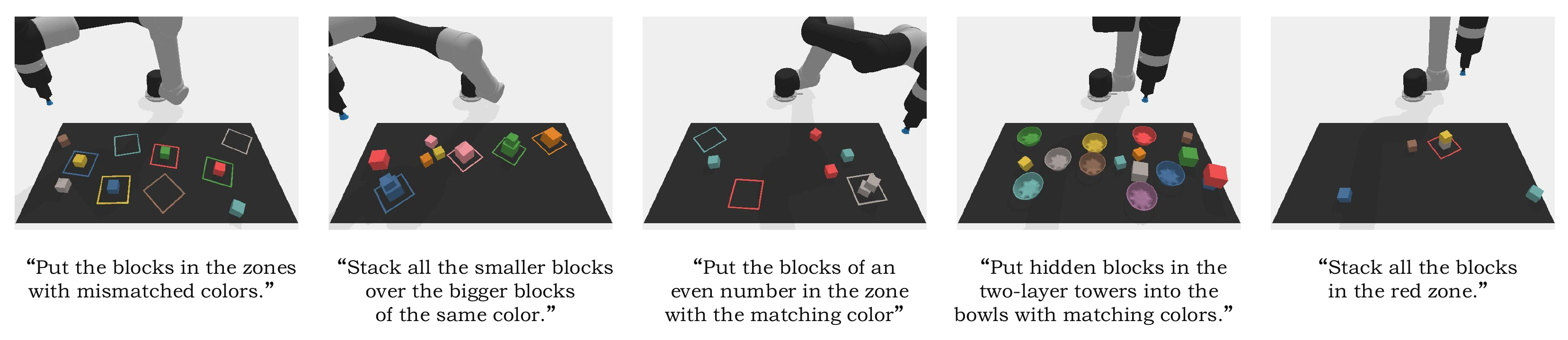}
\caption{An example of the long-horizon LoHoSet. 
Object attributes like size, color, quantity, and position vary across cases. 
}
\label{fig:dataset}
\end{figure*}

The training of LoHoVLA relies on a collection of demonstrations $(g, \gamma_t)$ with $\gamma_t = (\mathbf{o}_t, \hat{g}_t, \mathbf{a}_t)$.
% \mathbf{a}_t \mid \mathbf{o}_t, g, \hat{g}_t) \cdot \pi_\theta(\hat{g}_t \mid \mathbf{o}_t, g
% The policy $\pi_\theta$ can be tuned on a collection of such demonstrations are used to supervise the training of t. $\{g, \gamma_1, \gamma_2, \cdots ,\gamma_N\}$
The primary challenge is that sub-task annotation for real-world long-horizon tasks can rarely be obtained in a scalable manner without human intervention. 
To address this, we opt for a simulator-based approach following prior works~\cite{shridhar2022cliport, zhang2023lohoravens, meng2025data}.
Concretely, we construct the LoHoSet dataset based on the Ravens robot simulator~\cite{zeng2021transporter}. 
The simulation environment includes a UR5e robotic arm with a suction gripper and several objects placed on a table. 
The environment provides a reward signal only when the predicted action is both semantically correct and successfully executed. 
To simulate real-world uncertainties, the simulator adds observation noise and introduces a dropping probability $p$ for the end-effector to drop the picked block every second.
The visual observation $\mathbf{o} = (\mathcal{I}_{\text{color}}, \mathcal{I}_{\text{depth}})$ comprises RGB and depth top-down orthographic reconstructions of the scene. 
The goal instruction $g$ mainly focuses on rearranging the objects into the desired configuration (e.g., \textit{``Stack blocks in alternate colors on the green zone''}). %, the robot is tasked with rearranging the objects into the desired configuration.

We collect decomposed sub-tasks based on manually designed rules, thanks to the availability of complete information about the scenes from the simulator. 
In particular, we randomly initialize the scene, and we directly estimate the action $\mathbf{a} = (\mathcal{T}_{\text{pick}}, \mathcal{T}_{\text{place}})$ based on the positions of target objects. 
For sub-tasks without dependency constraints, their execution order is randomized; otherwise, we generate them with pre-defined logics. 
Each object is assigned a textual label, which is inserted into sub-task templates to generate sub-task descriptions (e.g., \textit{``Pick up the green block and place it in the green zone''} $\rightarrow$ \textit{``Pick up the blue block and place it on the green block''} $\rightarrow$ etc.). 
This process enables the generation of a large and diverse set of demonstrations.

% \zj{We how to decompose each goal as xxx}Each goal can be decomposed into a sequence of pick-and-place primitive sub-tasks (e.g., \textit{``Pick up the green block and place it in the green zone''} $\rightarrow$ \textit{``Pick up the blue block and place it on the green block''} $\rightarrow$ etc.). 
% We then \zj{how to obtain and collect} the atomic operation $\mathbf{a} = (\mathcal{T}_{\text{pick}}, \mathcal{T}_{\text{place}})$ corresponding to each sub-task $\hat{g}$.  

The resultant LoHoSet finally includes three types of objects: blocks, bowls, and zones, available in 11 distinct colors. 
The blocks are of two sizes, large and small. %, for generality. 
% The tasks are designed to evaluate long-horizon reasoning capabilities based on object attributes such as size, color, quantity, and spatial position. 
Following LoHoRavens~\cite{zhang2023lohoravens}, we adopt the 10 long-horizon tasks (detailed in Table~\ref{tab:taskid}) and 3 pick-and-place primitive tasks to facilitate comparison with the baselines.
We include 10 additional long-horizon tasks in LoHoSet to improve the generalization capacities of the trained model (see Figure~\ref{fig:evaluation} left). 
% \zj{rephrase this into the above}Moreover, we modify the scripts to generate sub-task labels for each long-horizon task, enabling training and evaluation of the model's task planning abilities. 
An example of these tasks is shown in Figure \ref{fig:dataset}. For more details on LoHoSet, please refer to Appendix A.

% \begin{figure*}[t]
% \centering
% % \hspace*{-0.16\textwidth}
% \includegraphics[width=1.0\textwidth]{dataset.pdf}
% \caption{An example of the long-horizon LoHoSet. 
% Object attributes like size, color, quantity, and position vary across cases. 
% }
% \label{fig:dataset}
% \end{figure*}
\subsection{Model Architecture}
\label{3.3}
% Below is a paragraph from a NeurIPS paper. Polish the writing to meet the academic style, improve the spelling, grammar, clarity, concision and overall readability. Simplify lengthy or cumbersome sentences. When necessary, rewrite the whole sentence. Firstly, you should provide the polished paragraph. Secondly, you should list all your modification and explain the reasons to do so in markdown table.

LoHoVLA employs a large pretrained vision language model (VLM) as its backbone to generate the next tokens, with specialized de-tokenizers to translate them as linguistic sub-tasks and actions, respectively. To address errors in both high-level planning and low-level control, it incorporates a hierarchical closed-loop control mechanism.
An overview of LoHoVLA is presented in Figure \ref{fig:architecture}.

\textbf{The Base Vision Language Model.} We select PaliGemma~\cite{beyer2024paligemma} as the backbone for our model due to its proven effectiveness in prior studies~\cite{black2024pi_0, intelligence2025pi_}. PaliGemma is a multi-modal foundation model processing both images and textual prompts for text generation. It integrates a SigLIP-based image encoder, a Gemma-2B decoder-only language model, and a linear projection layer that maps visual features into the language model’s token space.

% We retain the language head for sub-task planning and add an action head for robot action prediction.

\textbf{The Action De-Tokenizer.} Following prior work~\cite{ahn2022can, brohan2023rt, kim2024openvla}, we represent robot actions as discrete tokens to enable joint training with vision–language datasets. Specifically, we discretize the normalized action values into 1,024 uniform bins. During inference, robot actions are recovered by de-tokenizing and de-normalization.

\textbf{Hierarchical closed-loop control mechanism.} 
Compared to closed-loop control for atomic tasks, managing long-horizon tasks is more complex. Execution failures may arise from sub-task planning errors, inaccurate action predictions, or external disturbances. 
More formally, the three error types are:
(1) sub-task planning error,
(2) correct sub-task planning but incorrect action prediction, and
(3) correct planning and prediction, with failures caused by external disturbances.

% To address these challenges, we implement 
LoHoVLA embraces a hierarchical closed-loop control strategy that re-predicts actions more frequently than it re-plans sub-tasks. Specifically, if the number of failures during the current sub-task exceeds a predefined threshold $K$, the system triggers sub-task re-planning; otherwise, it only re-predicts the action. The control procedure used during evaluation is outlined in Algorithm~\ref{algorithm}. We assume that the robot receives a positive reward $r$ upon completing a sub-task. The $done$ flag is used to determine whether the overall goal has been achieved.

\begin{algorithm}[t]
\small
\caption{LoHoVLA Test-time Closed-Loop Control}
\label{algorithm}
\KwData{LoHoVLA policy $\pi_\theta$, initial observation $\mathbf{o}_0$, high-level goal $g$, reward $r$, \\ failure count $k$, failure threshold $K$, done flag $done$}
$t \gets 0$, $k \gets 0$, $r \gets 0$, $done \gets false$

\While{not $done$}{
\If{$t = 0$ or $r > 0$ or $k > K$}{
$\hat{g}_t \sim \pi_\theta(\hat{g}_t \mid \mathbf{o}_t, g)$\;

$k \gets 0$\;
}
$\mathbf{a}_t \sim \pi_\theta(\mathbf{a}_t \mid \mathbf{o}_t, g, \hat{g}_t)$\;

execute $\mathbf{a}_t$\;

get $\mathbf{o}_{t+1}, r, done$ from environment\;

\If{$r = 0$}{
$k \gets$ $k + 1$\;
}

$t \gets t + 1$;
}
\end{algorithm}

This design avoids unnecessary sub-task re-planning in cases (2) and (3). Our experiments demonstrate that the hierarchical closed-loop control scheme effectively mitigates errors in long-horizon tasks while avoiding redundant inference steps.

% \begin{algorithm}
% \caption{LoHoVLA Test-time Closed-Loop Control}
% \textbf{Require:} LoHoVLA policy $\pi_\theta$, initial observation $\mathbf{o}_0$, high-level goal $g$, reward $r$, \\ failure count $k$, failure threshold $K$, done flag $done$
% \begin{algorithmic}
% \STATE $t \gets 0, \, k \gets 0, \, r \gets 0, \, done \gets false$
% \WHILE{not $done$}
%     \IF{$t = 0$ \OR $r > 0$ \OR $k > K$}
%         \STATE $\hat{g}_t \sim \pi_\theta(\hat{g}_t \mid \mathbf{o}_t, g)$
%         \STATE $k \gets 0$
%     \ENDIF
%     \STATE $\mathbf{a}_t \sim \pi_\theta(\mathbf{a}_t \mid \mathbf{o}_t, g, \hat{g}_t)$
%     \STATE execute $\mathbf{a}_t$
%     \STATE $\mathbf{o}_{t+1}, \, r, \, done \gets$ environment
%     \IF{$r = 0$} 
%         \STATE $k \gets$ $k + 1$ 
%     \ENDIF
%     \STATE $t \gets t + 1$
% \ENDWHILE
% \end{algorithmic}
% \end{algorithm}
\subsection{Training Configurations}
\label{3.4}

During LoHoVLA’s training, we optimize the language model backbone while keeping the image encoder and the linear projection layer fixed. The training objective consists of two components: sub-task generation and action prediction. Both outputs are produced by the language model head and optimized using cross-entropy loss. The total loss is defined as:
\begin{equation}
\label{eq5}
    \mathcal{L} = \mathcal{L}_\text{text} + \mathcal{L}_\text{action}. 
\end{equation}
We adopt a two-stage training strategy. In the first stage, we fine-tune PaliGemma on long-horizon tasks, optimizing only the text loss to improve high-level task planning. In the second stage, we augment the dataset with pick-and-place primitive tasks and optimize both text and action losses to enhance action prediction.

\begin{table}[]
\caption{The seen and unseen tasks from LoHoRavens benchmark. Note that we also add the pick-and-place primitive task to the seen tasks.}
\label{tab:taskid}
\renewcommand{\arraystretch}{1.2}
\centering
\scriptsize
\begin{tabular}{@{}cll|l@{}}
\toprule[0.5mm]
\multicolumn{3}{c|}{\textbf{Tasks}}                                                                                                                                       & \multicolumn{1}{c}{\textbf{Instruction}}                                                                    \\ \midrule 
\multicolumn{1}{c|}{\multirow{5}{*}{\begin{tabular}[c]{@{}c@{}}Seen \\ Tasks\end{tabular}}}  & A. & pick-and-place-primitive                                     & ``Put the {[}OBJ{]} on the {[}OBJ / POS{]}.''                                                            \\
\multicolumn{1}{c|}{}                                                                        & B. & put-block-into-matching-bowl                                 & ``Put the blocks in the bowls with matching colors.''                                              \\
\multicolumn{1}{c|}{}                                                                        & C. & stack-smaller-over-bigger-with-same-color                    & ``Stack smaller blocks over bigger blocks of the same color.''                                     \\
\multicolumn{1}{c|}{}                                                                        & D. & stack-block-in-absolute-area                                 & ``Stack all the blocks in the {[}ABS POS{]} area.''                                                \\
\multicolumn{1}{c|}{}                                                                        & E. & put-even-blocks-in-same-color-zone                           & ``Move all blocks of a color that occur in even numbers.''                                         \\ \midrule \midrule
\multicolumn{1}{c|}{\multirow{6}{*}{\begin{tabular}[c]{@{}c@{}}Unseen\\ Tasks\end{tabular}}} & F. & put-block-into-mismatching-bowl                              & ``Put the blocks in the bowls with mismatching colors.''                                           \\
\multicolumn{1}{c|}{}                                                                        & G. & stack-blocks-of-same-size                                    & ``Stack blocks of the same size.''                                                                 \\
\multicolumn{1}{c|}{}                                                                        & H. & stack-blocks-with-alternate-color                            & ``Stack blocks in alternate colors.''                                                              \\
%\multicolumn{1}{c|}{} & \multirow{2}{*}{I.} & \makecell[tl]{stack-smaller-over-bigger-with-same-color- \\ in-same-color-zone} & \makecell[tl]{``Stack blocks of the same color in the zone with same color, \\ \, with the bigger blocks underneath.''} \\
%\multicolumn{1}{c|}{} &                      &                                                                                   &                                                                                              \\
\multicolumn{1}{c|}{} & \multirow{2}{*}{I.} & \makecell[tl]{stack-smaller-over-bigger-with-same-color- \\ in-same-color-zone} & \makecell[tl]{``Stack blocks of the same color in the zone with same color, \\ \, with the bigger blocks underneath.''} \\
\multicolumn{1}{c|}{}                                                                        & J. & move-blocks-between-absolute-positions                       & ``Move all the blocks in the {[}ABS POS{]} area to the {[}ABS POS{]} area.''                       \\
\multicolumn{1}{c|}{}                                                                        & K. & stack-blocks-of-same-size                                    & ``Stack blocks of the same color.''                                                                \\ \bottomrule[0.5mm]
\end{tabular}
\end{table}

\section{Experiments}

Our experimental evaluations aim to assess LoHoVLA’s capabilities in high-level task planning, low-level motion control, and generalization to novel tasks unseen during training. Specifically, we address the following questions:
\begin{itemize}[left=2pt, itemsep=0pt, parsep=4pt]
\item {How does LoHoVLA perform compared to hierarchical architecture baselines and standard VLA models in terms of performance and generalization on long-horizon tasks?}
\item {How effective are hierarchical closed-loop control schemes at mitigating errors in long-horizon tasks, where failures can arise from both high-level task planning and low-level motion control?}
\item {Can dataset expansion and a two-stage training strategy improve overall model performance?}
\end{itemize}

\subsection{Experimental Setup}
\label{sec:exset}
\textbf{Baselines.} We use LoHoRavens~\cite{zhang2023lohoravens} as the baseline with a hierarchical architecture. It consists of a Planner for high-level sub-task planning, an Actor for low-level motion control, and a Reporter for feedback-driven closed-loop control.  
It supports two types of Reporter-based feedback: (1) Explicit feedback, where the Planner (LLaMA 2 13B~\cite{touvron2023llama}) generates sub-tasks, the Actor (CLIPort~\cite{shridhar2022cliport}) executes them, and the Reporter (OpenFlamingo~\cite{awadalla2023openflamingo}) provides outcome captions to refine future plans; and (2) Implicit feedback, where the Reporter encodes visual observations using a frozen CLIP model~\cite{radford2021learning}, projects them via an MLP, and sends the embeddings to the Planner through LLaVA~\cite{liu2023visual} for continuous plan adjustment.
% LoHoRavens incorporates two types of Reporter-based feedback mechanisms: (1) Explicit feedback via captioning. The Planner (LLaMA 2 13B) generates sub-tasks, which are executed by the Actor (CLIPort). The Reporter (OpenFlamingo) produces natural language descriptions of the environment and action outcomes. These captions are fed back to the Planner to refine subsequent plans. (2) Implicit feedback via token projection. The Planner receives feedback in the form of token embeddings. The Actor executes the planned actions, while the Reporter encodes visual observations using a frozen CLIP model. These embeddings are projected via a single-layer MLP and passed to the Planner through the LLaVA interface, enabling continuous plan adjustment.

As a secondary baseline, we train a vanilla VLA model to isolate the effect of explicit sub-task prediction. This model directly predicts low-level robot actions without producing intermediate language outputs, relying solely on implicit sub-task inference. Our goal is to test whether it can learn sub-task reasoning without explicit planning.

\textbf{Training Details.}
LoHoVLA uses the PaliGemma-3b-mix-224 model as its backbone and is trained in two stages. In the first stage, we fine-tune PaliGemma on 14 long-horizon tasks—comprising 4 seen tasks from LoHoRavens and 10 additional tasks we designed—each with 1,000 demonstrations. We optimize only the text loss to improve high-level task planning. 
Unlike LoHoRavens, which uses prompt engineering to preserve generalization, our model requires fine-tuning, which leads to overfitting on the limited set of 4 tasks. To address this, we include 10 additional tasks to enhance generalization.
In the second stage, we augment the dataset with 10,000 demonstrations for each pick-and-place primitive and optimize both text and action losses to improve low-level motion control. 

Training is performed using the AdamW optimizer~\cite{loshchilov2017decoupled} with weight decay and gradient clipping. We utilize DeepSpeed~\cite{rasley2020deepspeed} for efficient distributed training. 
In training stage one, we train with 8 NVIDIA 4090 GPUs (24GB VRAM), a per-device batch size of 2, a learning rate of 5e-5, and for 3 epochs. We apply LoRA~\cite{hu2022lora} with a rank of 16 targeting all linear layers. 
In training stage two, we use a learning rate of 1e-5 for 1 epoch, keeping other settings unchanged. For comparison, we train the standard VLA model on the same dataset without sub-task labels using a learning rate of 1e-5, for 5 epochs, under the same hardware and LoRA configuration.

\textbf{Evaluation Metrics.} There are two evaluation methods for determining whether object states match the ground-truth, depending on the task category. The first is pose match, which requires the object's position and orientation to exactly align with the ground-truth. The second is zone match, where the overlap area between the predicted and ground-truth object must exceed a predefined threshold.
We evaluate each task instance using a score from 0 (failure) to 100 (success), based on the proportion of correctly completed pick-and-place steps. For example, if a task requires ten steps and the model completes eight, the score would be 80\%. In our evaluations, we report both the average score and the success rate on the test set.

\begin{table}[]
\caption{
% LoHoravens benchmark experimental results. For each task, we report the average reward (\%) and success rate (\%) of 200 episodes. LoHoVLA achieves the best or competitive performance across all tasks compared to baseline approaches. The bold entries correspond to highest success rates while underlined entries correspond to second-highest.
%For each task, we report the average reward (\%) and success rate (\%) over 200 episodes. LoHoVLA achieves the highest or comparable performance across all tasks relative to baseline models. Bold entries indicate the highest success rates; underlined entries indicate the second-highest.
Comparison of the average award (\%) and success rate (\%) on LoHoRavens benchmark. Bold entries indicate the highest success rates, underlined entries indicate the second-highest.
}
\renewcommand{\arraystretch}{1.2}
\centering
\small
\label{tab:mainresult}
\begin{tabular}{@{}cl|cccc@{}}
\toprule[0.5mm]
\multicolumn{2}{c|}{\multirow{2}{*}{\textbf{Tasks}}}                                                      & \multirow{2}{*}{\makecell[tl]{\; \;\;\;\textbf{Vanilla VLA} \; \;\;\;}} & \multicolumn{2}{c}{\textbf{LoHoRavens}}        & \multirow{2}{*}{\makecell[tl]{\; \;\;\; \textbf{LoHoVLA} \; \;\;\;}} \\ \cmidrule(lr){4-5}
\multicolumn{2}{c|}{}                                                                            &                              & Explicit feedback & Implicit feedback &                         \\ \midrule 
\multicolumn{1}{c|}{\multirow{5}{*}{\begin{tabular}[c]{@{}c@{}}Seen \\ Tasks\end{tabular}}}  & A & \textbf{79.0} / 79.0                  & 67.3 / -          & 67.3 / -          & \underline{77.5} / 77.5             \\
\multicolumn{1}{c|}{}                                                                        & B & 14.9 / 0.0                   & 31.4 / -          & \underline{37.0} / -          & \textbf{97.8} / 91.5             \\
\multicolumn{1}{c|}{}                                                                        & C & \underline{26.8} / 0.5                   & 18.0 / -          & 22.1 / -          & \textbf{34.9} / 22.5             \\
\multicolumn{1}{c|}{}                                                                        & D & 32.3 / 3.0                   & 30.4 / -          & \underline{33.2} / -          & \textbf{35.8} / 11.5             \\
\multicolumn{1}{c|}{}                                                                        & E & \underline{22.1} / 3.5                   & 9.6 / -           & 8.2 / -           & \textbf{85.1} / 81.0             \\ \midrule \midrule
\multicolumn{1}{c|}{\multirow{6}{*}{\begin{tabular}[c]{@{}c@{}}Unseen\\ Tasks\end{tabular}}} & F & \underline{52.1} / 9.0                   & 28.5 / -          & 21.1 / -          & \textbf{86.1} / 41.0             \\
\multicolumn{1}{c|}{}                                                                        & G & 6.8 / 0.0                    & \underline{21.9} / -          & 14.7 / -          & \textbf{40.1} / 25.0             \\
\multicolumn{1}{c|}{}                                                                        & H & 7.3 / 0.0                    & \underline{13.2} / -          & 5.2 / -           & \textbf{16.7} / 7.5              \\
\multicolumn{1}{c|}{}                                                                        & I & \underline{43.1} / 1.5                   & 12.8 / -          & 11.7 / -          & \textbf{77.2} / 52.0             \\
\multicolumn{1}{c|}{}                                                                        & J & \underline{38.6} / 10.5                  & 27.4 / -          & 27.2 / -          & \textbf{43.6} / 22.0             \\
\multicolumn{1}{c|}{}                                                                        & K & \underline{58.2} / 33.0                  & 4.0 / -           & 6.8 / -           & \textbf{73.8} / 54.5              \\ \bottomrule[0.5mm]
\end{tabular}
\end{table}

\subsection{Main Results}
% We evaluate LoHoVLA against the Vanilla VLA and LoHoRavens baselines on a benchmark suite comprising both seen and unseen tasks from LoHoRavens (Table ~\ref{tab:taskid}). LoHoVLA and Vanilla VLA are trained on five seen tasks and ten additional tasks we introduced. Evaluation results, including those of LoHoRavens with both explicit and implicit feedback mechanisms, are reported in Table~\ref{tab:mainresult}.

We evaluate LoHoVLA against the vanilla VLA and LoHoRavens on both seen and unseen tasks from LoHoRavens (Table~\ref{tab:taskid}). LoHoVLA and Vanilla VLA are trained on five seen tasks and ten additional tasks introduced in this work. Evaluation results, including those for LoHoRavens with both explicit and implicit feedback mechanisms, are shown in Table~\ref{tab:mainresult}.

LoHoVLA achieves the highest average score and success rate across nearly all tasks. On the \textit{put-block-into-matching-bowl} task, it attains near-perfect accuracy. 
On the most challenging reasoning task \textit{put-even-blocks-in-same-color-zone}, which requires integrating color recognition, counting, spatial reasoning, and logic, LoHoVLA achieves a score of 85.1 and a success rate of 81.0, while all baselines perform poorly.
Notably, LoHoVLA demonstrates strong generalization to unseen tasks, consistently outperforming all baselines despite having no prior exposure.

Interestingly, LoHoVLA occasionally outperforms on long-horizon tasks compared to \textit{pick-and-place-primitive} tasks. This is largely due to differences in evaluation criteria: zone-match tasks (e.g., involving bowls or colored zones) tolerate minor spatial inaccuracies, which LoHoVLA handles effectively. In contrast, pose-match tasks (e.g., block stacking) require precise alignment, where occasional sub-optimal motor trajectories can slightly reduce performance. Nevertheless, LoHoVLA remains robust across both task types.

Vanilla VLA performs the worst among all models, with zero success rates on several tasks. Our qualitative analysis reveals that this is primarily due to the absence of sub-task supervision, which leads the model to overfit to frequent patterns in the training data. For instance, in the \textit{put-block-into-matching-bowl} task, it often places blocks in incorrect bowls, disregarding the goal condition.

\subsection{Closed-Loop Strategies Comparison}
% \begin{table}[h]
% \small
% \centering
% \caption{\textbf{LoHoRavens Tasks}}

% \begin{tabular}{c|c|c}
% \toprule
% \multirow{5}{*}{\textbf{Seen tasks}} &
% A. Pick-and-Place primitives \\
% & B. ``Put the blocks in the bowls with matching colors'' \\
% & C. ``Stack smaller blocks over bigger blocks of the same color'' \\
% & D. ``Stack all the blocks in the [ABS\_POS] area'' \\
% & E. ``Move all blocks of a color that occur in even numbers \\
% & \quad to the same colored zone'' \\
% \midrule
% \multirow{6}{*}{\textbf{Unseen tasks}} &
% F. ``Put the blocks in the bowls with mismatching colors'' \\
% & G. ``Stack blocks of the same size'' \\
% & H. ``Stack blocks in alternate colors'' \\
% & I. ``Stack blocks of the same color in the zone with same color, \\
% & \quad with the bigger blocks underneath'' \\
% & J. ``Move all the blocks in the [ABS\_POS] area to the [ABS\_POS] area'' \\
% & K. ``Stack blocks of the same color'' \\
% \bottomrule
% \end{tabular}
% \end{table}

% \usepackage{booktabs}

% \usepackage{booktabs}
% \usepackage{multirow}
% \usepackage{booktabs}
% \usepackage{multirow}

% \usepackage{booktabs}
% \usepackage{multirow}

% \usepackage{booktabs}
% \usepackage{multirow}
\begin{table}[]
\caption{
Comparison of the average reward (\%), success rate (\%), and number of sub-task planning across three strategies.
Strategy (a) performs the worst. Strategies (b) and (c) are comparable overall, but (c) requires fewer high-level planning steps.
}
\renewcommand{\arraystretch}{1.2}
\centering
\small
\label{tab:closedloop}

\begin{tabular}{@{}cc|cc|cc|cc@{}}
\toprule[0.5mm]
\multicolumn{2}{c|}{\multirow{2}{*}{\makecell[tl]{\textbf{Tasks}}}}                                                      & \multicolumn{2}{c|}{\multirow{2}{*}{\makecell[tl]{ \;\;\;\; \textbf{Strategy (a)}\;\;\;\;}}}           & \multicolumn{2}{c|}{\multirow{2}{*}{\makecell[tl]{ \;\;\;\; \textbf{Strategy (b)}\;\;\;\;}}} & \multicolumn{2}{c}{\multirow{2}{*}{\makecell[tl]{ \;\;\;\; \textbf{Strategy (c)}\;\;\;\;}}} \\
\multicolumn{2}{c|}{}                                                                            & \multicolumn{2}{c|}{}                                        & \multicolumn{2}{c|}{}                              & \multicolumn{2}{c}{}                              \\ \midrule
\multicolumn{1}{c|}{\multirow{5}{*}{\begin{tabular}[c]{@{}c@{}}Seen \\ Tasks\end{tabular}}}  & A & 77.5 / 77.5                                            & 1.3 & 77.5 / 77.5                  & 1.3                 & 77.5 / 77.5                 & 1.3                 \\
\multicolumn{1}{c|}{}                                                                        & B & 89.5 / 74.0                                            & 5.7 & 96.4 / 88.5                  & 6.4                 & 97.8 / 91.5                 & 6.2                 \\
\multicolumn{1}{c|}{}                                                                        & C & 30.3 / 8.5                                             & 3.7 & 35.1 / 24.0                  & 7.9                 & 34.9 / 22.5                 & 7.8                 \\
\multicolumn{1}{c|}{}                                                                        & D & 20.1 / 2.5                                             & 2.2 & 31.3 / 9.0                   & 7.6                 & 35.8 / 11.5                 & 7.3                 \\
\multicolumn{1}{c|}{}                                                                        & E & 63.1 / 59.0                                            & 5.9 & 86.4 / 84.0                  & 6.6                 & 85.1 / 81.0                 & 6.2                 \\ \midrule \midrule
\multicolumn{1}{c|}{\multirow{6}{*}{\begin{tabular}[c]{@{}c@{}}Unseen\\ Tasks\end{tabular}}} & F & 62.3 / 29.0                                            & 3.6 & 86.2 / 42.0                  & 6.8                 & 86.1 / 41.0                 & 6.7                 \\
\multicolumn{1}{c|}{}                                                                        & G & 32.1 / 19.5                                            & 5.2 & 39.9 / 24.5                  & 8.2                 & 40.1 / 25.0                 & 7.9                 \\
\multicolumn{1}{c|}{}                                                                        & H & 14.1 / 4.0                                             & 4.5 & 15.9 / 6.0                   & 8.1                 & 16.7 / 7.5                  & 6.4                 \\
\multicolumn{1}{c|}{}                                                                        & I & 66.4 / 41.0                                            & 3.7 & 78.2 / 56.0                  & 6.9                 & 77.2 / 52.0                 & 6.8                 \\
\multicolumn{1}{c|}{}                                                                        & J & 31.3 / 14.5                                            & 4.2 & 44.5 / 23.5                  & 7.4                 & 43.6 / 22.0                 & 7.1                 \\
\multicolumn{1}{c|}{}                                                                        & K & 52.9 / 34.0 & 4.7 & 71.1 / 52.0                  & 7.2                 & 73.8 / 54.5                 & 6.9                 \\ \bottomrule[0.5mm]
\end{tabular}
\end{table}

% \begin{tabular}[c]{@{}c@{}}52.9 \\ / 34.0\end{tabular}
To assess the effectiveness of our specialized closed-loop control mechanism in addressing task execution failures, we compare the following three control strategies.
\begin{itemize}[left=2pt, itemsep=0pt, parsep=4pt]
\item {
\textbf{Strategy $\,$(a).} On failure, the system re-predicts the action only without re-planning the sub-task. 
}
\item{
\textbf{Strategy $\,$(b).} The system re-plans the sub-task and then re-predicts the action after every failure. 
}
\item{
\textbf{Strategy (c).} The hierarchical closed-loop control strategy: if the number of failures within the current sub-task exceeds a predefined threshold $K$, the system initiates sub-task re-planning; otherwise, it only re-predicts the action. In our experiments, we set $K=2$.
}
\end{itemize}

In addition to measuring the average reward and success rate, we track the number of high-level sub-task planning. All three strategies are evaluated using the fine-tuned LoHoVLA model. Results are reported in Table~\ref{tab:closedloop}.

As expected, strategy (a) yields the poorest performance. When failures stem from incorrect sub-task planning, this approach continues executing flawed plans, potentially leading to deadlocks. Strategies (b) and (c) perform comparably in overall metrics; however, strategy (c) requires fewer high-level sub-task planning steps. This is because many failures result from low-level prediction errors or external disturbances, where re-planning the sub-task is unnecessary.

Furthermore, task characteristics influence strategy performance. For reasoning-heavy tasks (e.g., \textit{put-even-blocks-in-same-color-zone}), strategy (b) excels by promptly correcting sub-task assignments. In contrast, for tasks demanding precise motor control (e.g., \textit{stack-block-in-absolute-area}), strategy (c) performs better, as repeated low-level attempts improve the likelihood of success.

\begin{figure}[htbp]
  \centering
  \subfigure[Evaluate Traing Set Expansion]{
    \includegraphics[width=0.6666\textwidth]{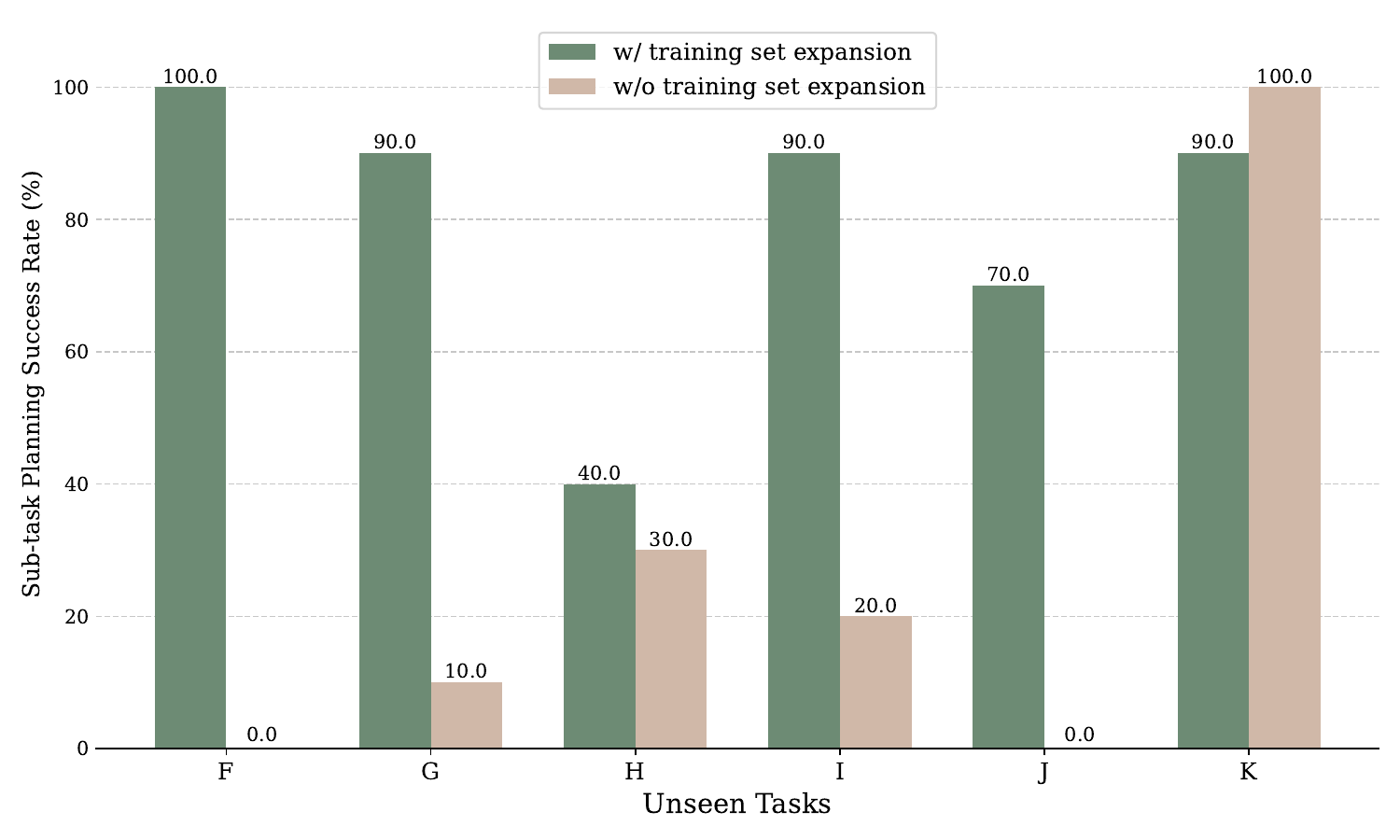}
  }
  \subfigure[Evaluate Two-stage Training]{
    \includegraphics[width=0.297\textwidth]{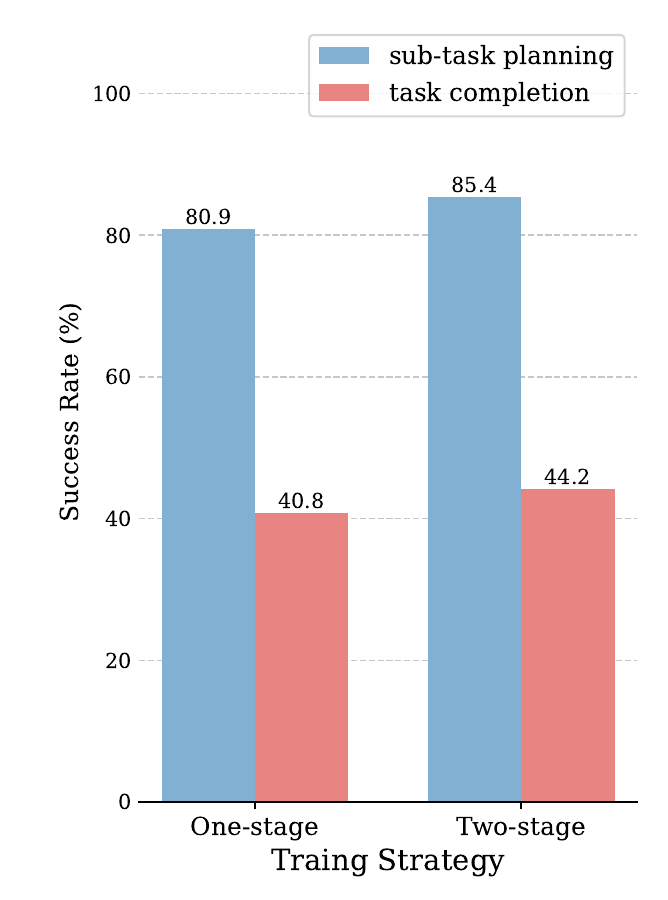}
  }
  \caption{
(a) Comparison of performance on unseen tasks between training with and without dataset expansion, evaluated using the sub-task planning success rate (\%). (b) Comparison of performance between one-stage and two-stage training strategies, evaluated using both the sub-task planning success rate (\%) and task completion success rate (\%).
  }
  \label{fig:evaluation}
\end{figure}

\subsection{Ablation Studies}
We examine the effects of training set expansion and the two-stage training method on model performance. To better evaluate planning ability, we introduce a new metric: sub-task planning success rate. Unlike sub-task execution success, this metric isolates planning performance by removing the influence of low-level action prediction and external disturbances. Specifically, for each task, we randomly sample 10 timesteps from all episodes and manually enumerate all valid subsequent sub-tasks. An LLM is then used to assess whether the model’s predicted sub-task at each timestep is semantically equivalent to any of the enumerated ground-truth options. Further details are provided in Appendix B.

\textbf{Evaluate Training Set Expansion.} We trained LoHoVLA with and without training set expansion (i.e., the extra 10 tasks used for training mentioned in Section~\ref{3.2}), and evaluated the generalization ability on the unseen tasks of the LoHoRavens Benchmark. The results are shown in Figure~\ref{fig:evaluation} left. It can be observed that the model trained without extra data has poor generalization ability. 
This is due to severe overfitting to the seen tasks. For example, the success rate of the task \textit{put-block-into-mismatching-bowl} is 0, because its scene is similar to that of put-block-into-matching-bowl, which led the model to overfit to the latter and ignore the language goal, placing the block into the matching-colored bowl instead. The expanded dataset alleviates such overfitting issues.

\textbf{Evaluate Two-stage Training.} We conduct both one-stage and two-stage training experiments on LoHoVLA using identical configurations. In both settings, the model is trained for 5 epochs. The key difference is that in the two-stage approach, the primitive tasks and action labels are introduced only after the first 3 epochs. The results—average sub-task planning success rate (\%) and average task success rate (\%)—are presented in Figure~\ref{fig:evaluation} right. The one-stage training setup yields a lower sub-task planning success rate, which consequently results in reduced task success. This indicates that introducing action labels and primitive tasks too early hinders the effective optimization of high-level task planning.

\section{Conclusion}
\label{sec:conclu}
For long-horizon embodied tasks requiring both high-level planning and low-level control, existing VLA models and hierarchical approaches struggle with planning and coordination. 
To address this, we propose LoHoVLA, a unified VLA framework that uses a large pretrained vision-language model to jointly generate sub-tasks and robot actions. 
It incorporates a hierarchical closed-loop control mechanism to correct errors at both levels. 
Empirical results show that LoHoVLA outperforms prior VLA and hierarchical methods by decent margins and demonstrates strong generalization. % and advancing embodied intelligence.

\textbf{Limitations.}
The limitations stem primarily from the limited precision of robot actions due to their discrete structure. Additionally, our assumption that a sub-task can be completed within a single timestep may not be practical in real-time applications.

% \newpage
%\bibliographystyle{plain}  % 或者neurips_2025默认的样式
%\bibliography{neurips_2025}
% \printbibliography
\newpage
\bibliographystyle{plainnat}
\bibliography{neurips_2025}

%%%%%%%%%%%%%%%%%%%%%%%%%%%%%%%%%%%%%%%%%%%%%%%%%%%%%%%%%%%%

\newpage
\appendix
\section{LoHoSet}
\label{sec:append}
% \input{tabs/alltasks}

% Below is a paragraph from a NeurIPS paper. Polish the writing to meet the academic style, improve the spelling, grammar, clarity, concision and overall readability. Simplify lengthy or cumbersome sentences. When necessary, rewrite the whole sentence. Firstly, you should provide the polished paragraph. 

\begin{figure*}[h]
\centering
\includegraphics[width=1.0\textwidth]{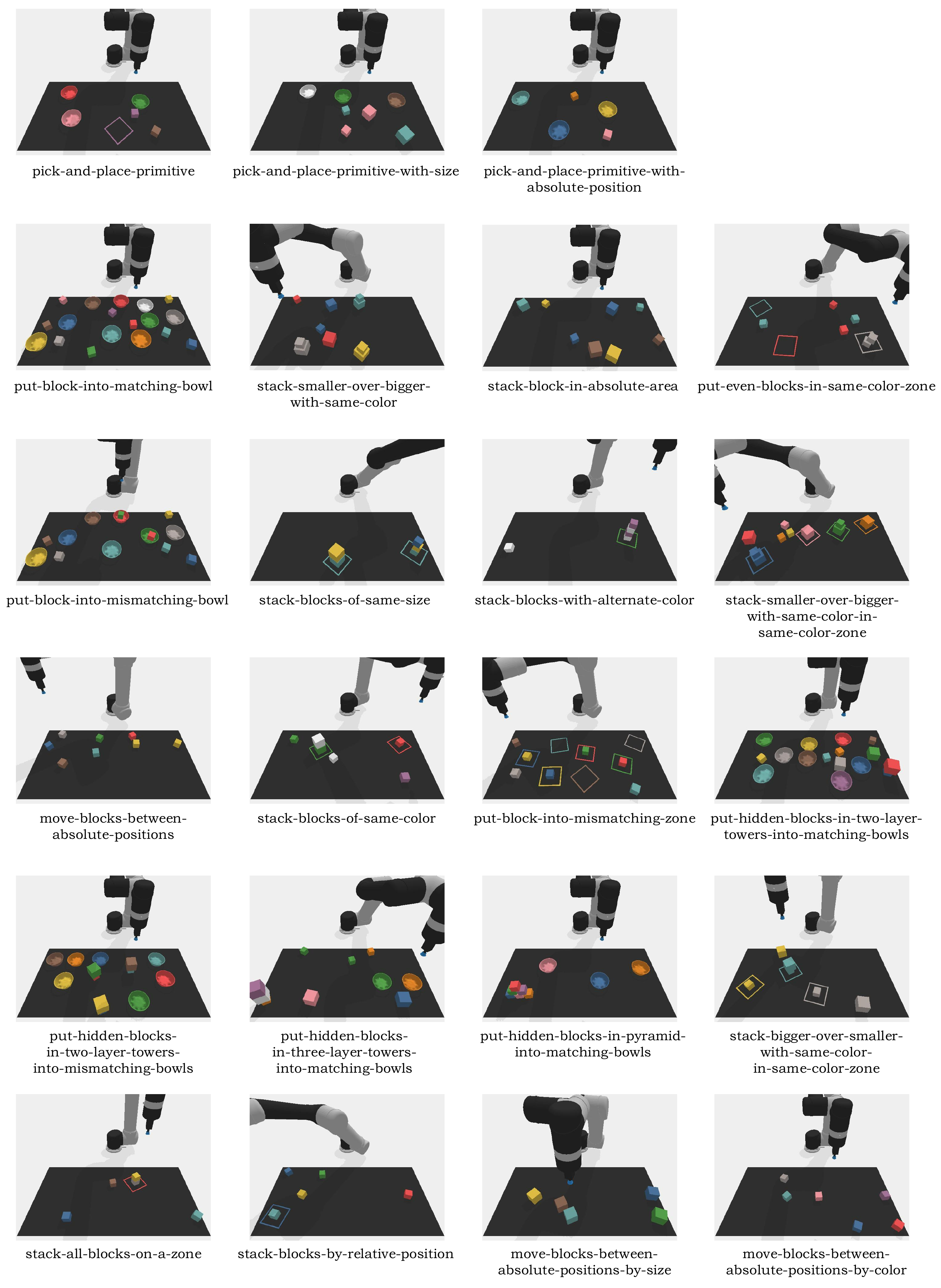}
\caption{Visual examples of all tasks in the LoHoSet dataset, including 3 pick-and-place primitives and 20 long-horizon tasks, showcasing the diversity and complexity of the task scenarios.
}
\label{fig:dataset_complete}
\end{figure*}

\newpage

LoHoSet consists of 3 pick-and-place primitive tasks and 20 long-horizon tasks. Among them, 10 long-horizon tasks and all 3 primitive tasks are adapted from the LoHoRavens benchmark to facilitate comparison with existing baselines. The other 10 long-horizon tasks are introduced to improve the generalization ability of the trained model. A complete list of tasks is provided in Table~\ref{tab:alltasks}, and examples of all tasks are illustrated in Figure~\ref{fig:dataset_complete}.

\begin{table}[h]
\caption{Summary of all tasks in LoHoSet, including 3 primitive pick-and-place tasks and 20 long-horizon tasks, with distinctions between those adopted from LoHoRavens and those introduced for improved generalization.}
\label{tab:alltasks}
\renewcommand{\arraystretch}{1.2}
\centering
\scriptsize
\begin{tabular}{@{}cl|l@{}}
\toprule[0.5mm]
\multicolumn{2}{c|}{\textbf{Tasks}}                                                                                                                                                                                        & \multicolumn{1}{c}{\textbf{Instruction}}                                                                                                                    \\ \midrule
\multicolumn{1}{c|}{\multirow{7}{*}{\begin{tabular}[c]{@{}c@{}}LoHoRavens\\ Seen Tasks\end{tabular}}}   & pick-and-place-primitive                                                                                & “Put the {[}OBJ{]} on the {[}OBJ{]}.”                                                                                                              \\
\multicolumn{1}{c|}{}                                                                                   & pick-and-place-primitive-with-size                                                                      & “Put the {[}SIZE{]} {[}OBJ{]} on the {[}SIZE{]} {[}OBJ{]}.”                                                                                          \\
\multicolumn{1}{c|}{}                                                                                   & pick-and-place-primitive-with-absolute-position                                                         & “Put the {[}OBJ{]} on the {[}ABS\_POS{]}.”                                                                                                         \\
\multicolumn{1}{c|}{}                                                                                   & put-block-into-matching-bowl                                                                            & “Put the blocks in the bowls with matching colors.”                                                                                                \\
\multicolumn{1}{c|}{}                                                                                   & stack-smaller-over-bigger-with-same-color                                                               & “Stack smaller blocks over bigger blocks of the same color.”                                                                                       \\
\multicolumn{1}{c|}{}                                                                                   & stack-block-in-absolute-area                                                                            & “Stack all the blocks in the {[}ABS\_POS{]} area.”                                                                                                  \\
\multicolumn{1}{c|}{}                                                                                   & put-even-blocks-in-same-color-zone                                                                      & “Move all blocks of a color that occur in even numbers.”                                                                                           \\ \midrule \midrule
\multicolumn{1}{c|}{\multirow{6}{*}{\begin{tabular}[c]{@{}c@{}}LoHoRavens\\ Unseen Tasks\end{tabular}}} & put-block-into-mismatching-bowl                                                                         & “Put the blocks in the bowls with mismatching colors.”                                                                                             \\
\multicolumn{1}{c|}{}                                                                                   & stack-blocks-of-same-size                                                                               & “Stack blocks of the same size.”                                                                                                                   \\
\multicolumn{1}{c|}{}                                                                                   & stack-blocks-with-alternate-color                                                                       & “Stack blocks in alternate colors.”                                                                                                                \\
\multicolumn{1}{c|}{}                               
& 
% \begin{tabular}[c]{@{}l@{}} 
\makecell[tl]{stack-smaller-over-bigger-with-same-color-\\ in-same-color-zone} 
% \end{tabular} 

& 
% \begin{tabular}[c]{@{}l@{}}
\makecell[tl]{“Stack blocks of the same color in the zone with same color,\\    \, with the bigger blocks underneath.” }
% \end{tabular}     
\\

% \multicolumn{1}{c|}{} & \multirow{2}{*}{I.} & \makecell[tl]{stack-smaller-over-bigger-with-same-color- \\ in-same-color-zone} & \makecell[tl]{``Stack blocks of the same color in the zone with same color, \\ \, with the bigger blocks underneath.''} \\
% \multicolumn{1}{c|}{}

\multicolumn{1}{c|}{}                                                                                   & move-blocks-between-absolute-positions                                                                  & “Move all the blocks in {[}ABS\_POS{]} area to {[}ABS\_POS{]} area.”                                                                       \\
\multicolumn{1}{c|}{}                                                                                   & stack-blocks-of-same-color                                                                               & “Stack blocks of the same color.”                                                                                                                  \\ \midrule \midrule
\multicolumn{1}{c|}{\multirow{10}{*}{\begin{tabular}[c]{@{}c@{}}Additional\\ Tasks\end{tabular}}}       & put-block-into-mismatching-zone                                                                         & “Put the blocks in the zones with mismatching colors.”                                                                                             \\
\multicolumn{1}{c|}{}                                                                                   & 

% \begin{tabular}[c]{@{}l@{}}put-hidden-blocks-in-two-layer-towers-\\ into-matching-bowls\end{tabular}    & \begin{tabular}[c]{@{}l@{}}“Put all the hidden blocks in the two-layer stacked towers \\     into the bowls with matching colors.”\end{tabular}

\makecell[tl]{put-hidden-blocks-in-two-layer-towers-\\ into-matching-bowls}
& 
\makecell[tl]{“Put all the hidden blocks in the two-layer stacked towers \\  \,   into the bowls with matching colors.”}

\\
\multicolumn{1}{c|}{}                                                                                   & 

% \begin{tabular}[c]{@{}l@{}}put-hidden-blocks-in-two-layer-towers-\\ into-mismatching-bowls\end{tabular} & \begin{tabular}[c]{@{}l@{}}“Put all the hidden blocks in the two-layer stacked towers \\     into the bowls with mismatching colors.”\end{tabular} 
\makecell[tl]{put-hidden-blocks-in-two-layer-towers-\\ into-mismatching-bowls}
& 
\makecell[tl]{“Put all the hidden blocks in the two-layer stacked towers \\  \,   into the bowls with mismatching colors.”}

\\
\multicolumn{1}{c|}{}                                                                                   & 

\makecell[tl]{put-hidden-blocks-in-three-layer-towers-\\ into-matching-bowls}  & 
\makecell[tl]{“Put all the hidden blocks in the three-layer stacked towers \\  \,   into the bowls with matching colors.”}  \\
\multicolumn{1}{c|}{}                                                                                   & 

\multirow{2}{*}{\makecell[tl]{put-hidden-blocks-in-pyramid-\\ into-matching-bowls}}            & 
\makecell[tl]{“Put all the hidden blocks on the first layer of the pyramid \\  \,   into the bowls with matching colors.”}  \\
\multicolumn{1}{c|}{}                                                                                   & 

\multirow{2}{*}{\makecell[tl]{stack-bigger-over-smaller-with-same-color-\\ in-same-color-zone}} & 
\makecell[tl]{“Stack blocks of the same color in the zone with same color,\\   \,  with the smaller blocks underneath.”}    \\
\multicolumn{1}{c|}{}                                                                                   & stack-all-blocks-on-a-zone                                                                              & “Stack all the blocks on the {[}COLOR{]} zone.”                                                                                                    \\
\multicolumn{1}{c|}{}           &                                                                        \multirow{2}{*}{stack-blocks-by-relative-position}                                                                       & 
\makecell[tl]{“Stack all the blocks on the {[}REL\_POS{]} of the {[}COLOR{]} block \\ \, on the {[}COLOR{]} zone.”}                                                     \\
\multicolumn{1}{c|}{}       &                                                                            \multirow{2}{*}{move-blocks-between-absolute-positions-by-size}                                                          & \makecell[tl]{“Move all the {[}SIZE{]} blocks in the {[}ABS\_POS{]} area \\ \, to the {[}ABS\_POS{]} area.”}                                                            \\
\multicolumn{1}{c|}{}                                                                                   & \multirow{2}{*}{move-blocks-between-absolute-positions-by-color}                                                         & \makecell[tl]{“Move all the {[}COLOR{]} blocks in the {[}ABS\_POS{]} area \\ \, to the {[}ABS\_POS{]} area.”}                                                           \\ \bottomrule[0.5mm]
\end{tabular}
\end{table}

% \multicolumn{1}{c|}{} & \multirow{2}{*}{I.} & \makecell[tl]{stack-smaller-over-bigger-with-same-color- \\ in-same-color-zone} & \makecell[tl]{``Stack blocks of the same color in the zone with same color, \\ \, with the bigger blocks underneath.''} \\
% \multicolumn{1}{c|}{}    

\newpage

\section{Ablation Studies Details}

We investigate the effects of training set expansion and a two-stage training strategy on model performance. To more accurately evaluate planning capabilities, we introduce a new metric: sub-task planning success rate. Unlike sub-task execution success, this metric isolates high-level planning by excluding the influence of low-level action prediction and external noise. For each task, we randomly sample 10 timesteps across all episodes and manually enumerate valid next sub-tasks. A large language model (LLM) then assesses whether the model’s predicted sub-task at each timestep is semantically equivalent to any of the ground-truth options. The evaluation prompt is as follows:

\vspace{1em}
\begin{tcolorbox}[colback=gray!5!white, colframe=gray!75!black, title=Evaluation Prompt]
You are given the current sub-task in a sequential task and a predicted next sub-task generated by a model. 
Additionally, you are provided with a list of valid ground-truth next sub-tasks. 
Determine whether the predicted sub-task is semantically equivalent to any of the ground-truth options. 
Focus solely on semantic similarity in intent and meaning, ignoring differences in wording or phrasing. 
Respond with 'Yes' if the prediction is semantically equivalent to at least one ground-truth sub-task; otherwise, respond with 'No'.
\end{tcolorbox}

\end{document}